# A submodular-supermodular procedure with applications to discriminative structure learning

**Mukund Narasimhan**
Department of Electrical Engineering
University of Washington
Seattle WA 98004
mukundn@ee.washington.edu

**Jeff Bilmes**
Department of Electrical Engineering
University of Washington
Seattle WA 98004
bilmes@ee.washington.edu

## Abstract

In this paper, we present an algorithm for minimizing the difference between two submodular functions using a variational framework which is based on (an extension of) the concave-convex procedure [17]. Because several commonly used metrics in machine learning, like mutual information and conditional mutual information, are submodular, the problem of minimizing the difference of two submodular problems arises naturally in many machine learning applications. Two such applications are learning discriminatively structured graphical models and feature selection under computational complexity constraints. A commonly used metric for measuring discriminative capacity is the EAR measure which is the difference between two conditional mutual information terms. Feature selection taking complexity considerations into account also fall into this framework because both the information that a set of features provide and the cost of computing and using the features can be modeled as submodular functions. This problem is NP-hard, and we give a polynomial time heuristic for it. We also present results on synthetic data to show that classifiers based on discriminative graphical models using this algorithm can significantly outperform classifiers based on generative graphical models.

## 1 Introduction

The importance of submodularity in machine learning and data mining applications has been demonstrated in [13, 14, 22]. These papers apply submodularity to tasks such as PAC learning structure of graphical models and to finding a set of nodes in a social network that maximize influence while satisfying a cardinality constraint. These papers exploit the fact that entropy, mutual information, and the influence of a set of nodes in a social network are submodular set functions. Intuitively, submodularity corresponds to the notion of diminishing returns (a formal definition will be given in Section 2). So, for example, one would expect (intuitively) that the information capacity of subsets of features could be formulated as a submodular function. In many cases, even when the function of interest is not submodular, but is *approximately* submodular, algorithms for submodular optimization can be relied on to behave reasonably well [13]. Many machine learning/data mining applications use entropy or (conditional) mutual information as metrics for selecting features and/or structure in these models. For example many common algorithms for learning decision trees use mutual information to select attributes for internal nodes of the trees. Similarly, many algorithms for feature selection [16], and determining capacity of multiaccess fading channels [15], can all be formulated as solutions to standard submodular optimization problems. This also includes the algorithm proposed by Chow and Liu [1] for learning trees because finding the minimum spanning tree algorithm can be regarded as the problem of finding a maximum weight basis in a matroid (as the rank function of a matroid is a submodular function).

There are many different optimization problems that arise in these applications. The algorithm presented in [13] requires computing a set that minimizes a submodular set function. A polynomial time algorithm for this problem was first presented in [3]. However, this algorithm was useless for all practical purposes because it was based on the ellipsoid algorithm. A pseudo polynomial algorithm for minimizing a submodular set function was given by Cunningham [6], while Schrijver [12] and Iwata, Fleischer and Fujishige [9] gave fully polynomial algorithms for this problem, which made the problem somewhat practical for small instances. Queyranne [5] presented an algorithm for this problem when the set function also satisfies a symmetry condition, which only requires $O(n^3)$ time (where $n$ is the size of the underlying ground set), making the problem feasible for medium to large instances.

The algorithm presented in [14] requires computing a set that is an approximate maximizer of a submodular set function given a cardinality constraint. This problem is clearly NP-hard because it contains NP-complete problems such as MAX-CUT as special cases. Lovasz [4] has shown that in general, this problem can take exponential time independent of the P vs. NP question. However it was shown in [7] that a greedy algorithm finds an approximate solution that is guaranteed to be within $\frac{e-1}{e} \sim 0.69$ of the optimal solution. Several other problems, such as selecting features to maximize the total information content of the features can be seen to be special cases of this problem.

The algorithm used in [15] for computing a rate vector that satisfies capacity constraints requires computing a modular function that is bounded above by a submodular function. This is because the capacity region of a multiaccess channel is bounded by submodular functions. It is shown in [3] that this problem can be solved optimally by a greedy algorithm.

In this paper, we propose another optimization problem, namely finding a set that minimizes the difference between two submodular functions. This generalizes the problem of maximizing a submodular function, and hence this problem is NP-hard as well. There are at least two very interesting applications that can be formulated as a solution to this problem. The first is the problem of selecting a subset of features given a complexity constraint (as opposed to a cardinality constraint). The second is the problem of learning discriminatively structured graphical models. We present some results on synthetic examples for the second.

## 2 Preliminaries

**Definition 1.** *A function $f : 2^V \to \mathbb{R}$ is submodular if for every $A, B \subseteq V$, we have*

$$f(A) + f(B) \geq f(A \cup B) + f(A \cap B)$$

Two common examples of submodular functions are given below.

**Lemma 2.** *Suppose $\{X_v\}_{v \in V}$ is a collection of random variables, and $f : 2^V \to \mathbb{R}$ is given by $f(S) = H(\{X_v\}_{v \in S})$, the entropy of the collection of random variables $\{X_v\}_{v \in V}$. Then $f$ is a submodular function.*

*Proof.* $0 \leq I(A; B) = H(A) + H(B) - H(A \cup B) - H(A \cap B)$. □

Similarly, if $S$ is an arbitrary collection of random variables, then the function $f : 2^V \to \mathbb{R}$ given by $f(A) = H\left(\{X_v\}_{v \in A} \mid \{X_v\}_{v \in S}\right)$ is also submodular.

**Lemma 3.** *Suppose $\{X_v\}_{v \in V}$ is a collection of random variables, and $f : 2^V \to \mathbb{R}$ is given by $f(A) = I\left(\{X_v\}_{v \in A}; \{X_v\}_{v \in V \setminus A}\right)$, the mutual information function. Then $f$ is a submodular function.*

*Proof.* We can write $f(A) = H(A) + H(V \setminus A) - H(V)$, where $H(X)$ is the binary entropy of the set of random variables $X$. Therefore, $f(A) + f(B) = [H(A) + H(B)] + [H(V \setminus A) + H(V \setminus B)] - 2H(V)$. By the submodularity of $H(\cdot)$, $H(A) + H(B) \geq H(A \cap B) + H(A \cup B)$. Similarly, $H(V \setminus A) + H(V \setminus B) \geq H(V \setminus (A \cup B)) + H(V \setminus (A \cap B))$. The result now follows by noting that $f(A \cap B) = H(A \cap B) + H(V \setminus (A \cap B)) - H(V)$ and $f(A \cup B) = H(A \cup B) + H(V \setminus (A \cup B)) - H(V)$. Therefore $f(\cdot)$ is submodular. □

The class of submodularity is preserved under several naturally occurring operations. For example, scaling by positive constants, and adding submodular functions preserves submodularity. Now, if $f$ is submodular, and we think of $f(A)$ as the "gain" associated with using the set $A$, then submodularity corresponds to the notion of diminishing returns. More specifically, if we define the incremental gain at $A$ to be

$$\rho_f(A, x) = f(A \cup \{x\}) - f(A)$$

Then $f$ is submodular if and only if $\rho_f$ is nonincreasing (so $A \subseteq B$ implies that $\rho_f(A, x) \geq \rho_f(B, x)$ for every $x \notin B$). When $x \in V$ and $A \subseteq V$, we will use $A + x$ to denote the set $A \cup \{x\}$. A function $g$ is supermodular if $-g$ is submodular. A function is modular if it is both submodular and supermodular. In particular, if $f$ is submodular, and $h$ is modular, then $-h$ is submodular, and so $f - h$ is submodular. Every modular function $h$ can be written as $h(A) = \sum_{x \in A} h(\{x\})$. We can therefore associate a modular function $h$ with a vector in $\mathbb{R}^{|V|}$, namely the vector $\langle h(\{x\}) \rangle_{x \in V}$. Given a submodular function $f$, the extended polymatroid associated with $f$ is the subset of $\mathbb{R}^{|V|}$ given by

$$E_f = \left\{ h \in \mathbb{R}^{|V|} : h(S) \leq f(S) \, \forall S \subseteq V \right\}$$

So, every element $h \in E_f$ is a modular lower bound on $f$.

Now, given that this paper is about minimizing the sum of a submodular and a supermodular function, one might ask how broadly applicable this theory is. In fact, set functions that can be expressed as the sum of a submodular and a supermodular function are ubiquitous as the following result shows.

**Lemma 4.** *Every set function $\phi$ can be expressed as the sum of a submodular and supermodular function.*

*Proof.* The proof of this follows from a result of Lovasz [4] in which it is shown that a set function is submodular (resp. supermodular) if and only if its so called Lovasz extension is convex (resp. concave). Let $f$ be a submodular function whose Lovasz extension $\hat{f}$ has a Hessian with all eigenvalues bounded below by $\epsilon > 0$. Let $\hat{\phi}$ be the Lovasz extension of $\phi$. Now, it can be verified that the Lovasz extension of the sum of functions is the sum of the Lovasz extensions of the individual functions. For sufficiently large $\lambda > 0$, the Lovasz extension $\widehat{\phi + \lambda \cdot f}$ of $\phi + \lambda \cdot f$ has positive definite Hessian. Therefore, $\phi + \lambda \cdot f$ is submodular. Therefore, $\phi$ is the sum of a submodular function $\phi + \lambda \cdot f$, and a supermodular function $-\lambda \cdot f$. □

In the following sections, we describe two machine learning applications in which the difference of submodular functions arises more naturally.

### 2.1 Feature Selection

Suppose that $F$ is a collection of features. Associated with each (sub)set of features, we have a measure of the information contained in the features $g : 2^V \to \mathbb{R}$. If each of the features is a random variable, then a reasonable choice for $g$ is the (joint) entropy of the collection of features. We will only require that $g$ be a submodular function, which is a very reasonable assumption (and is true for the entropy function). Suppose that there is some computational cost for computing (or using) each feature. In general, it is not true that the cost of computing the features are equal and independent. That is, it is often the case that the cost of computing a collection of features (jointly) is less the the cost of computing each feature individually. For example, if we use features based on the Fourier spectrum of a signal, then to compute the features we need to perform an FFT on the signal. However, once the FFT has been computed, we can reuse the computation for evaluating all the features that depend on the Fourier spectrum. Hence we have a certain "economy of scale" for feature computation. Therefore, if we use $c : 2^V \to \mathbb{R}$ as a measure of the complexity of computing each set of features, then it is reasonable to model $c$ as a submodular function. What we are looking for is a set of features that do not cost too much to compute, but are very informative. However, in general, these goals are somewhat contradictory, and so we seek a good tradeoff between the two functions. That is, we seek a "best" set of features which minimizes a function of the form $g - k \cdot c$, where $k \in \mathbb{R}^+$ is a constant. One could just constrain the number of allowable features, and the greedy algorithm gives a good approximation [7] (of factor $\frac{e-1}{e}$) for this problem. However, this is not necessarily the best way of proceeding since (a) not all features are equally complex to compute and (b) the submodularity of $c$ (the feature computation cost) ensures that the approximation could be arbitrarily bad. Finding the minimizer of $g - k \cdot c$ is a better approach as it yields the best tradeoff between the cost of computing/using the features and the potential gain obtained by using those features. Therefore, the problem of minimizing the difference of two submodular functions arises here naturally.

### 2.2 Discriminatively structured graphical models

When using graphical models in pattern classification tasks, it is is important to produce models that can classify well even if they do not necessarily represent in a generative fashion the patterns that they are trying to classify. In general there have been two approaches to designing classifiers: generative and discriminative. Generative classifiers use a generative model $p(X|C)$ where $X$ represents a set of features and $C$ is the class variable. The training procedure typically attempts to pick the parameters of the distribution $p_\theta(X|C)$ to maximize the likelihood of training data. Discriminative models on the other hand, pick parameters to maximize the ability of the model to discriminate between the classes [19]. It has been observed that generatively trained models are not ideal for classification. In the past, work has proceeded to trying to produce generative models that discriminate better. The TAN models for example ([18]) augment the structure of Naïve Bayes so that $p(X|C)$ is a tree. They do this by augmenting edges to either maximize the mutual information $I(X; Y)$, conditional mutual information $I(X; Y|C)$ or class-specific structures $I(X; Y|C = c)$. Each of these structure learning criterion is generative in nature and does not optimize classification directly. Another goal is to produce a generative model whose inherent factorization structure is optimized from a discriminative point of view [20].

Here is a simple example which shows that generative models can lead to very bad classification results. Suppose that there are two classes, corresponding to the (unobserved) random variable $C$, and three (observed) binary random variables $X_1, X_2, X_3$. For both classes, $X_1$ is uniformly distributed, and $X_2 = X_1$ with probability 0.5 and is a random coin toss with probability 0.5. For class 1, $X_3$ is generated according to the fol-

|  | $C = 1$ | | $C = 2$ | |
|---|---|---|---|---|
|  | $X_3 = 0$ | $X_3 = 1$ | $X_3 = 0$ | $X_3 = 1$ |
| $X_1 X_2 = 00$ | $\frac{21}{64}$ | $\frac{3}{64}$ | $\frac{15}{64}$ | $\frac{9}{64}$ |
| $X_1 X_2 = 01$ | $\frac{3}{64}$ | $\frac{5}{64}$ | $\frac{1}{64}$ | $\frac{7}{64}$ |
| $X_1 X_2 = 10$ | $\frac{5}{64}$ | $\frac{3}{64}$ | $\frac{7}{64}$ | $\frac{1}{64}$ |
| $X_1 X_2 = 11$ | $\frac{3}{64}$ | $\frac{21}{64}$ | $\frac{9}{64}$ | $\frac{15}{64}$ |

Table 1: The probability distributions for $C \in \{1, 2\}$

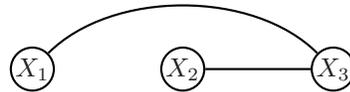

Figure 2: A discriminative structure.

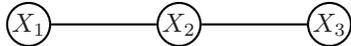

Figure 1: The optimal generative structure.

lowing:

$$X_3 = \begin{cases} X_2 & \text{with probability } 0.5 \\ X_1 & \text{with probability } 0.25 \\ \text{independent coin toss} & \text{with probability } 0.25 \end{cases}$$

For class 2, $X_3$ is given by

$$X_3 = \begin{cases} X_2 & \text{with probability } 0.5 \\ -X_1 & \text{with probability } 0.25 \\ \text{independent coin toss} & \text{with probability } 0.25 \end{cases}$$

The exact probabilities are shown in Table 1. To find the best tree graphical model, we can use the procedure proposed by Chow and Liu in [1]. However, since there are two classes, and hence two probability distributions, we could

1. Use the class specific mutual information $I(X; Y | C = c)$ to weight the edge $\{X, Y\}$ to build one tree structure for each class (for $X, Y$ scalars).

2. Use the mutual information $I(X; Y)$ to weight the edge $\{X, Y\}$ to build one common tree structure for both classes.

3. Use the condition mutual information $I(X; Y | C)$ to weight the edge $\{X, Y\}$ to build one common tree structure for both classes.

It can be verified that in all three cases, the graphical model results in the linear chain shown in Figure 1. Once the structure has been determined, the parameters of the model are chosen to minimize the KL-divergence between the true distribution and the reduced complexity model. Table 2 shows that even the optimal generative graphical model that meets the specified complexity constraints can yield bad results.

This is in part because the dependence of $X_3$ on $X_1$ is the only dependence that allows one to discriminate between the classes. However, this dependence is much weaker than the dependence between $X_2$ and $X_3$. Unfortunately, this dependence is identical across classes, and modeling this dependence at the expense of the $X_3/X_1$ dependence does not improve the classification performance. Therefore, using only the (conditional) mutual information between the random variables as the metric for selecting edges as is done in [1] can result in bad classification performance. Now it is intuitively clear, that the graphical model shown in Figure 2 models the distinction between classes. This is borne out by Table 2 which shows the (asymptotic) performance of classification using the exact probability distribution, the generative model shown in Figure 1 and the discriminative model shown in Figure 2. The results using the full model are the best we can possibly hope to do. In this case, Naïve Bayes offers no advantage over random guessing, and while the generative model does yield better results, the discriminative model performs much better. The key point here is that finding the optimal generative model using an algorithm like Chow-Liu does not give good performance because it is the solution to the "wrong" problem. This is true even on an extremely simple model. A commonly used metric which better approximates the desired discriminative capability of links is given by the EAR criterion $I(X; Y | C) - I(X; Y)$ (see [20]), which gives the optimal tree in this case. The EAR measure is an approximation of the expected log conditional likelihood (i.e. class posterior distribution). Both terms $I(X; Y | C)$ and $I(X; Y)$ are submodular when $Y = V \setminus X$, and hence optimizing the EAR criterion leads to minimizing the difference of submodular functions. The EAR criterion attempts to measure the degree to which the independence assumption $X \perp\!\!\!\perp Y | C$ in augmented Naïve Bayes would be a good assumption. Intuitively, this is because for classification, it is desirable to model structure between variables that are independent conditional on $C$ but not unconditionally independent. In the past, approaches have not even yielded locally optimal solutions.

|            | Complete | Generative | Discriminative |
|------------|----------|------------|----------------|
| Error Rate | 0.375    | 0.437      | 0.406          |

Table 2: Asymptotic classification error rate

## 3 The submodular-supermodular procedure

In this section, we describe the submodular-supermodular procedure, and discuss its relation to the convex-concave procedure and aforementioned applications. For this, we first describe the essential idea behind the concave-convex procedure [17] in a suitably abstract setting, and show that while one set of choices leads to the concave-convex procedure, another set of choices leads to the submodular-supermodular procedure.

Suppose that we are interested in minimizing a function $\phi : D \to \mathbb{R}$ over a given domain $D$, where $\phi = f+g$ can be expressed as the sum of a submodular function $f : D \to \mathbb{R}$ and a supermodular function $g : D \to \mathbb{R}$ (or alternatively as the difference of two submodular functions $f$ and $-g$). We do this by finding a sequence of elements $x_0, x_1, \ldots, x_n, \ldots$ in $D$ so that $\phi(x_0) \geq \phi(x_1) \geq \phi(x_2) \geq \ldots \phi(x_n)$ using the following procedure. We start by picking an arbitrary $x_0 \in D$. Suppose that we have already picked points $x_0, x_1, \ldots, x_n$. Now, find a function $h_n : D \to \mathbb{R}$ which satisfies

1. $h_n(x_n) = g(x_n)$.
2. $h_n(x) \geq g(x)$ for every $x \in D$.
3. $\phi_n(x) := f(x) + h_n(x)$ can be minimized efficiently.

This immediately implies that for any $x \in D$, we have

$$\phi_n(x) = f(x) + h_n(x) \geq f(x) + g(x) = \phi(x)$$

Therefore, $\phi_n$ is an upper-bound for $\phi$. This bound is tight at $x_n$ because $h_n(x_n) = g(x_n)$. Now, pick $x_{n+1} = \arg\min_{x \in D} \phi_n(x)$ We can do this efficiently because $\phi_n$ is "nice" for minimization. Then we get

$$\phi(x_{n+1}) \leq \phi_n(x_{n+1}) \leq \phi_n(x_n) = \phi(x_n)$$

Therefore, by induction, we can build up a sequence $x_0, x_1, \ldots, x_n, x_{n+1}, \ldots$ so that $\phi(x_0) \geq \phi(x_1) \geq \ldots \phi(x_n) \geq \phi(x_{n+1}) \geq \ldots$ This is a decreasing sequence, and so must have a limit (which could be $-\infty$). If $\phi$ is continuous and $D$ is compact, then we can get arbitrarily close to a local minimum. On the other hand if $D$ is a finite set, the sequence eventually becomes constant (achieving a local minima). The concave-convex procedure [17] is a specific instance of this procedure where $f$ and $g$ are convex and concave respectively. When $g : D \to \mathbb{R}$ is concave, we have for any $x, x_n \in D$,

$$g(x) \leq g(x_n) + \nabla_g(x_n) \cdot (x - x_n)$$

Therefore, if we define $h_n(x) := g(x_n) + \nabla_g(x_n) \cdot (x - x_n)$, then $h_n$ satisfies the conditions listed above. It is clear that $h_n : D \to \mathbb{R}$ is an affine function and so if $f(x) : D \to \mathbb{R}$ is a convex function, then $\phi_n = f + h$ is still a convex function, and so can be minimized efficiently. Note that if $x_{n+1} \in \arg\min_{x \in D} \phi_n(x)$, then we must have $\nabla_{\phi_n}(x_{n+1}) = 0$ (if $x$ is in the interior of $D$), and hence

$$0 = \nabla_{\phi_n}(x_{n+1}) = \nabla_f(x_{n+1}) + \nabla_{h_n}(x_{n+1})$$
$$= \nabla_f(x_{n+1}) + \nabla_g(x_n)$$

Hence $\nabla_f(x_{n+1}) = -\nabla_g(x_n)$, which is the iterative update rule for the concave-convex procedure.

It is generally accepted that the discrete analog of convex functions are the submodular functions and the discrete analog of concave functions are supermodular functions. Therefore, it is quite natural to expect that the procedure can also be applied for minimizing the sum of a submodular and supermodular functions, or alternatively, the difference of two submodular functions. More concretely, suppose that we have a ground set $V$, and we are interested in minimizing some function $\phi : 2^V \to \mathbb{R}$ over all subsets of $V$. Let $\phi(A) = f(A) - g(A)$, where $f, g : 2^V \to \mathbb{R}$ are submodular functions. Then to apply the procedure we need to be able to compute at any point $A_n \in 2^V$, the "modular approximation" to $g$ which is exact at $A_n$. More specifically, we seek a function $h_n : 2^V \to \mathbb{R}$ which satisfies

1. $h_n(A_n) = g(A_n)$
2. $h_n(A) \leq g(A)$ for all $A \in 2^V$
3. $\phi_n := f - h_n$ is submodular.

We therefore get

$$\phi_n(A_n) = f(A_n) - h_n(A_n) = f(A_n) - g(A_n) = \phi(A_n)$$

and for all $A \in 2^V$, we have

$$\phi_n(A) = f(A) - h_n(A) \geq f(A) - g(A) = \phi(A)$$

We pick $A_{n+1} = \arg\min_{A \in 2^V} \phi_n(A)$. We can do this efficiently because the sum of a modular and submodular functions is submodular, and minimizers of submodular functions can be computed in time polynomial in $|V|$. Therefore, we have

$$\phi(A_{n+1}) \leq \phi_n(A_{n+1}) \leq \phi_n(A_n) = \phi(A_n)$$

and we may inductively build up a sequence $A_0, A_1, A_2, \ldots, A_n, A_{n+1}, \ldots$ such that

$$\phi(A_0) \geq \phi(A_1) \geq \phi(A_2) \geq \cdots \geq \phi(A_n) \geq \phi(A_{n+1}) \geq \cdots$$

Since $2^V$ is finite, the process eventually terminates. We can guarantee different kinds of local minima at termination by picking the modular approximations differently.

In order to make this procedure into an algorithm, there are several details we need to fill in. First, we need to specify how the minimization of the submodular functions are carried out. Second, we need to demonstrate that "modular approximations" of submodular functions can be computed efficiently. Finally, we need to specify a termination condition.

### 3.1 Minimization of submodular functions

For a general submodular function, we can use either the algorithm due to Schrijver [12] or the algorithm by Iwata, Fleischer and Fujishige [9] to find a minimum. However, these algorithms, while polynomial in the size of the ground set $|V|$, are still very expensive ($O(|V|^8)$ and $O(|V|^7)$ respectively). If the given submodular function is symmetric as well (so for every $A \subseteq V$, we have $f(A) = f(V \setminus A)$), then we can use the algorithm due to Queyranne [5] to find a minimizer. Queyranne's algorithm run in time $O(|V|^3)$, and so is quite efficient. Recently, Nagamochi and Ibaraki [10] and Rizzi [11] have shown that Queyranne's algorithm in fact works for a larger class of submodular functions than just the symmetric ones. In [10], it is shown that this algorithm computes the correct minimizer if the function is posimodular. A posimodular function is a function that satisfies

$$f(A) + f(B) \geq f(A \setminus B) + f(B \setminus A)$$

Note that a symmetric submodular function is posimodular, and so the class of posimodular functions are a superset of the class of symmetric submodular functions. So, if $f - h_n$ is posimodular, then Queyranne's algorithm could be used to find its minimizer instead of a more general algorithm. Using a fast algorithm like Queyranne's algorithm instead of a more general algorithm can make a tremendous difference in the running times of reasonably sized instances (though the running time continuous to be polynomial).

### 3.2 Computing modular approximations

Given a submodular function $g : 2^V \to \mathbb{R}$, we seek a modular approximation $h_n : 2^V \to \mathbb{R}$ that is exact at $A_n$. That is, we need $h_n(A) \leq g(A)$ for all $A \subseteq V$ and $h_n(A_n) = g(A_n)$. Essentially, we seek an element $h$ that is contained in $Eg$, the extended base polymatroid of $g$. The following two results were first obtained by Edmonds [2].

**Proposition 5.** *Suppose that $g : 2^V \to \mathbb{R}$ is a submodular function, and $\pi$ is any permutation of the set $V$. Let $W_i = \{\pi(1), \pi(2), \ldots, \pi(i)\}$, so $W_{|V|} = V$. We define a function $h : V \to \mathbb{R}$ as follows.*

$$h(\pi(i)) = \begin{cases} g(W_1) & \text{if } i = 1 \\ g(W_i) - g(W_{i-1}) & \text{otherwise} \end{cases}$$

*This gives a function $h$ that is defined only on the single element subsets of $V$. It can however easily be extended to all subsets of $V$ by defining it to be*

$$h(A) = \sum_{x \in A} h(x)$$

*for every $A \subseteq V$. Then*

1. $h(A) \leq g(A)$ for every $A \subseteq V$.
2. $h(W_m) = g(W_m)$ for every $1 \leq m \leq |V|$.

*Proof.* See [3]. □

Observe that $h(W_i) = g(W_i)$ for $1 \leq i \leq |V|$.

---
**Algorithm 1:** submodularSupermodular

**Data**: Submodular function $g$ and $f$
**Result**: An approximate minimizer of $f - g$
**begin**
    $n \leftarrow 0$
    $\pi_0 \leftarrow$ random permutation of $V$
    improvementFound $\leftarrow$ true
    min $\leftarrow \infty$
    **while** improvementFound **do**
        $h_n \leftarrow$ modularApproximation$(g, \pi_n)$
        $A_n \leftarrow \arg\min_{A \subseteq V, A \neq \phi, V}(f - h_n)(A)$
        val $\leftarrow (f - h_n)(A_n)$
        $\pi_{n+1} \leftarrow$ random permutation starting with $A_n$
        $n \leftarrow n + 1$
        **if** val $<$ min $- \delta$ **then**
            min $\leftarrow$ val
            improvementFound $\leftarrow$ true
    **return** $A_{n-1}$
**end**

---

This gives us a procedure to compute a function $h$ that is modular and is bounded above by $g$. Since $h(W_i) = g(W_i)$ for every $1 \leq i \leq |V|$, if we order the elements of $V$ so that $A = W_{|A|}$, we we get a modular approximation with all the desired properties. We say that the permutation begins with $A$ (see Algorithm 1). Further, this modular approximation is the tightest possible approximation to $g$ in the following sense.

**Proposition 6.** *Every h obtained as above is a vertex in the extended base polymatroid of g, and every vertex in the extended base polymatroid of g can be obtained by picking an appropriate permutation. Moreover, if c is a vector in $\mathbb{R}^{|V|}$, then the greedy procedure described above is the result of the optimization problem*

$$\max c \cdot x \text{ subject to } x \in Eg$$

*Proof.* See [3]. □

---
**Algorithm 2:** modularApproximation

---
**Data**: A submodular function $g$ and a permutation $\pi$
**Result**: A modular approximation $h$
**begin**
    $W_0 \leftarrow \phi$
    **for** $i \leftarrow 1$ **to** $|V|$ **do**
        $W_i \leftarrow W_{i-1} \cup \{\pi(i)\}$
        $h(\pi(i)) \leftarrow g(W_i) - g(W_{i-1})$
**end**
**return** h

---

Algorithm 2 gives an algorithm that computes a modular approximation for a submodular function $g$ with the desired properties. This is used as a subroutine in the algorithm shown in Algorithm 1 which is a simple variational algorithm for minimizing the difference between two submodular functions. Finally, we note that we can modify Algorithm 1 slightly to guarantee different kinds of local optima. To see this note that the modular approximation $h$ for $g$ based on a permutation $\pi$ satisfies $h(W_i) = g(W_i)$ for $1 \leq i \leq |V|$ (using the terminology of Proposition 5). In particular if $\pi$ begins with $A_n$, then $W_k = A_n$ for $k = |A_n|$. In this case, $h$ is also exact for $W_{k-1} = A_n \setminus \{\pi(k-1)\}$ and for $W_{k+1} = A_n \cup \{\pi(k+1)\}$. Therefore, we know that if $A_n$ is the final solution produced, then $\phi(A_n) \leq \phi(A_n \setminus \{\pi(k-1)\})$ and $\phi(A_n) \leq \phi(A_n \cup \{\pi(k+1)\})$. Therefore, to guarantee that the current solution cannot be improved by either the removal or deletion of any single element, we only need to ensure that we consider the permutation in which that element is either $\pi(k+1)$ or $\pi(k-1)$. Therefore, by considering at most $O(n)$ permutations, we can guarantee that the current solution cannot be improved by the addition or deletion of any single element. Similarly, to guarantee that the current solution cannot be improved by either the removal or deletion of at most $k$ elements, we need only consider $O(n^k)$ permutations for computing the modular approximation and then picking the best. The $\delta$ parameter is used in the algorithm mainly to account for numerical precision effects, and the desired degree of approximation. It can be set to 0 if we are using a fully combinatorial algorithm for submodular minimization.

# 4 Learning discriminatively structured graphical models

In this section, we describe how the supermodular-submodular procedure described can be used to learn discriminatively structured trees. The algorithm we describe is a variant of the algorithm described in [13]. In [13], the optimal separators are identified by solving a series of submodular minimization problems, and then a dynamic programming formulation is used to build up the optimal model. In this work, we use a greedy strategy to identify the best separator, along with the components that the separator divides the graph into. This best separator is taken to be one that minimizes the EAR criterion. Then we recursively compute optimal models for the two components thus identified. While this procedure could be used to find graphical models of treewidth greater than 1, we only use it to identify 1-trees. This is because it can then be compared with the optimal generative tree generated using the algorithm of Chow and Liu [1].

The algorithm is shown in Algorithm 3.

---
**Algorithm 3:** makeDiscriminativeTree

---
min $\leftarrow \infty$
**for** $x \in V$ **do**
    $V_x \leftarrow V \setminus \{x\}$
    prtn $\leftarrow \arg \min_{S \neq \phi, V_x} I(S; V \setminus S | x, C) - I(S; V \setminus S | x)$
    val $\leftarrow I(\text{prtn}; V \setminus \text{prtn}|x) - I(\text{prtn}; V \setminus \text{prtn}|x, C)$
    **if** val $<$ min **then**
        min $\leftarrow$ val
        minSeparator $\leftarrow \{x\}$
        minPartition $\leftarrow$ prtn

root $\leftarrow$ minSeparator
makeDiscriminativeTree (separator $\cup$ minPartition)
makeDiscriminativeTree (separator $\cup$ $V \setminus$ minPartition)

---

# 5 Experimental Results

To test our algorithm, we generated synthetic data consisting of jointly distributed Gaussian random variables. Two distributions corresponding to two difference classes were generated so that there is a strong discriminative component. The random variables in the two classes have identical means and variances, so for example, a naïve Bayes classifier will be useless. Because the means and variances are identical, a classifier needs to model the "structure" in order to be effective. However, the data was generated so that there is a lot of dependence between the random variables that is very similar in both classes. The strength of the dependences that provide the discriminative capability is considerably less than the strength that is common to both classes. Because of this, the optimal generative tree will end up modeling structure that is not very useful for classification purposes. Table 3 shows the results of classification on 2000 randomly generated examples. In this example, the oracle was simulated by estimating the covariance matrix from another 2000 randomly generated samples. The same covariance matrix was used for finding both the generative and the discriminative trees. We used Schrijver's algorithm [12] to minimize the submodular functions. The results indicate that the discriminative tree does provide a measurable advantage over the generative tree. However, this is perhaps not very surprising as the data was deliberately generated to show the advantage of a discriminative model over a generative model. To see that this was better than picking a random tree, $|V|$ random trees were generated, and the average and best error rates are also reported. Naïve Bayes was also tested, but it did not provide a significant advantage over random guessing. These experiments do not prove that such a technique will work well on real world applications (we expect to test this in future work). We view these results more as a proof of the concept that the EAR measure can serve as a reasonable measure of discrimination capability, and that we can use the supermodular-submodular algorithm to find structure that is better that the generative model. Note that the algorithm we have presented is not guaranteed to find the optimal discriminative tree, and so it is possible that the optimal discriminative model can perform considerably better than the results shown.

## 6 Conclusions

While the results of this paper seem to indicate that the EAR measure is a reasonable approximation for the classifier performance, it is still just an approximation. The supermodular-submodular procedure does not produce a (globally) optimal solution, and so using this procedure results in an solution that approximately minimizes an objective function that approximates the desired criterion. We could then use a dynamic programming procedure similar to the one described in [13], but instead, we use a simple greedy strategy for recursively partitioning the variables and constructing a tree. Thus the results we present cannot really be considered to be the "optimal" discriminative solution, but they still outperform the optimal generative solution. Therefore, these results serve as a proof of the concept that we can find better classifiers using a generative model using the submodular-supermodular procedure (or possibly other algorithms).

There are clearly several questions we have left unanswered. First, it would be extremely desirable to show bounds on the performance of this algorithm, or alternatively, find another algorithm with guaranteed performance bounds. Second, we have given no bounds on the number of iterations required for convergence of the procedure. Ideally, we should be able to bound the number of iterations by some (small degree) polynomial in the number of variables. Empirically, we observe that the number of iterations required for the specific case when the submodular and supermodular functions are mutual information terms of jointly distributed guassians is small.

## 7 Acknowledgements

This work was supported by NSF grant IIS-0093430 and an Intel Corporation Grant.

| Number of of variables | Complete Model | Best Random Tree | Avg. Random Tree | Generative Tree | Discriminative Tree | Time per run(sec) |
|---|---|---|---|---|---|---|
| 5 | 0.362 | 0.464 | 0.469 | 0.476 | 0.455 | 0.3 |
| 6 | 0.392 | 0.445 | 0.461 | 0.452 | 0.373 | 0.4 |
| 7 | 0.186 | 0.427 | 0.450 | 0.401 | 0.367 | 0.4 |
| 8 | 0.246 | 0.441 | 0.458 | 0.446 | 0.358 | 0.6 |
| 9 | 0.189 | 0.416 | 0.435 | 0.404 | 0.389 | 1.2 |
| 10 | 0.047 | 0.429 | 0.453 | 0.420 | 0.365 | 2.1 |
| 15 | 0.004 | 0.433 | 0.464 | 0.447 | 0.399 | 8.4 |

Table 3: Error rates of various classifiers for synthetic data